\documentclass[10pt,twocolumn,letterpaper]{article}

\usepackage{wacv}
\usepackage{times}
\usepackage{epsfig}
\usepackage{graphicx}
\usepackage{amsmath}
\usepackage{amssymb}
\usepackage{tabularx}
\usepackage{multirow}
\usepackage{float}
\usepackage{booktabs}

\def\wacvPaperID{384} 

\wacvfinalcopy 
\ifwacvfinal
\usepackage[breaklinks=true,bookmarks=false]{hyperref}
\else
\usepackage[pagebackref=true,breaklinks=true,colorlinks,bookmarks=false]{hyperref}
\fi

\makeatletter
\newcommand{\printfnsymbol}[1]{%
  \textsuperscript{\@fnsymbol{#1}}%
}

\begin{document}

\title{\ RGPNet: A Real-Time General Purpose Semantic Segmentation}

\author{Elahe Arani\thanks{Equal contribution.}, Shabbir Marzban\footnotemark[1], Andrei Pata, and Bahram Zonooz\\
Advanced Research Lab, NavInfo Europe, Eindhoven, The Netherlands\\
{\tt\small \{elahe.arani, shabbir.marzban, andrei.pata\}@navinfo.eu, bahram.zonooz@gmail.com}
}

\maketitle

\begin{abstract}
 We propose a real-time general purpose semantic segmentation architecture, RGPNet, which achieves significant performance gain in complex environments. RGPNet consists of a light-weight asymmetric encoder-decoder and an adaptor. The adaptor helps preserve and refine the abstract concepts from multiple levels of distributed representations between encoder and decoder. It also facilitates the gradient flow from deeper layers to shallower layers. Our experiments demonstrate that RGPNet can generate segmentation results in real-time with comparable accuracy to the state-of-the-art non-real-time heavy models. Moreover, towards green AI, we show that using an optimized label-relaxation technique with progressive resizing can reduce the training time by up to 60\% while preserving the performance. We conclude that RGPNet obtains a better speed-accuracy trade-off across multiple datasets.
\end{abstract}

\section{Introduction}
\begin{figure*}[tb]
 \centering
 \includegraphics[width=.98\textwidth]{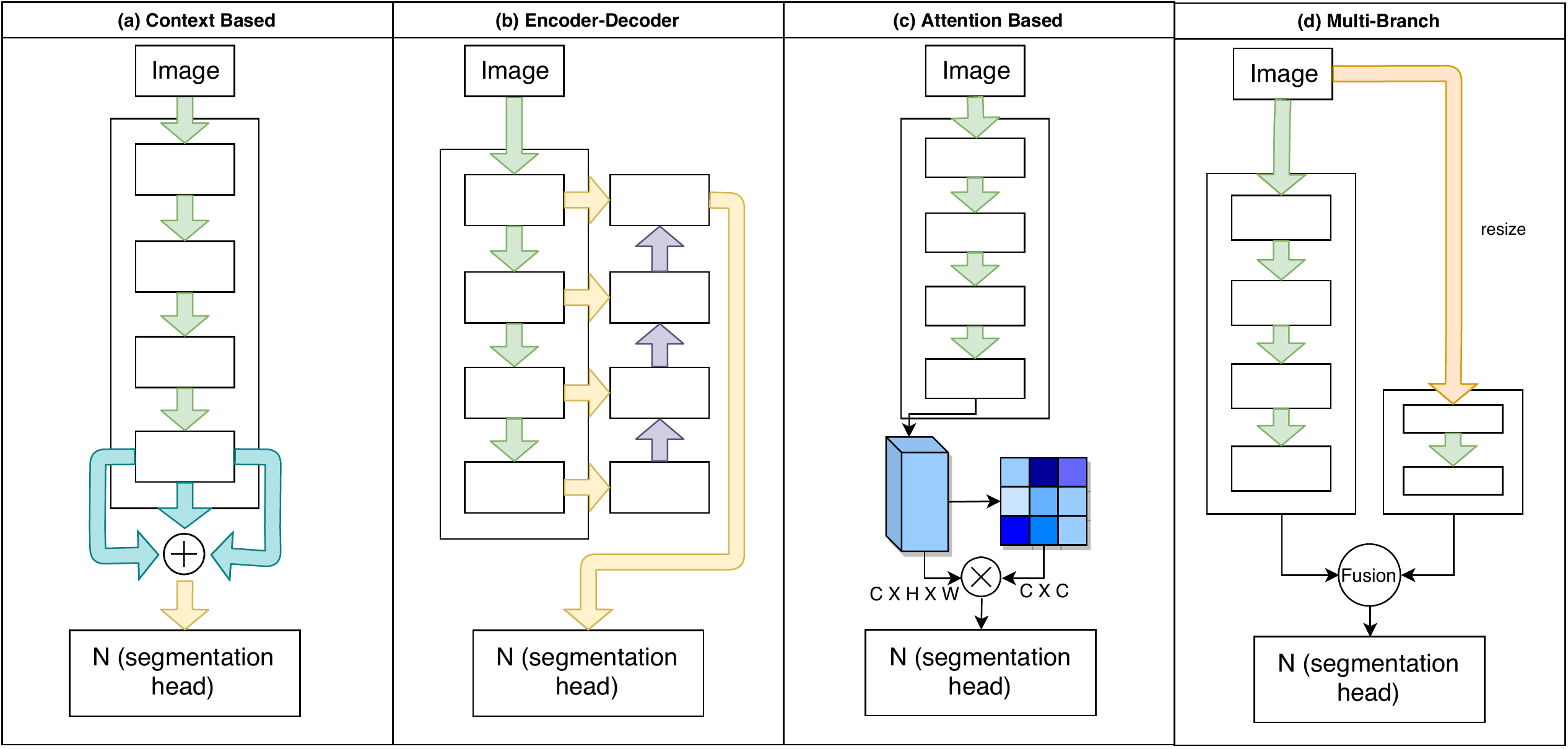}
 \caption{ Schematic illustrations of common semantic segmentation architectures. (a) In context-based networks, dilated convolutions with multiple dilation rates are employed in cascade or in parallel to capture a multi-scale context. (b) In encoder-decoder networks, encoder extracts the features of high-level semantic meaning and decoder densify the features learned by the encoder. (c) In attention-based networks, the feature at each position is selectively aggregated by a weighted sum of the features at all positions. This can be done across channels or spatial dimensions. (d) Multi-branch networks are employed to combine semantic segmentation results at multiple resolution levels. The lower resolution branches yield deeper features with reduced resolution and the higher resolution branches learn spatial details.} 
 
 \label{fig:lit_review}
\end{figure*}

Convolutional neural networks (CNNs) have brought about a paradigm shift in the field of computer vision, leading to tremendous advances in many tasks \cite{girshick2015fast,he2016deep,krizhevsky2012imagenet,lan2018person,li2017person,simonyan2014very,szegedy2015going}.
Semantic segmentation, which associates each pixel to the object class it belongs to, is a computationally expensive task in computer vision \cite{long2015fully}. Fast semantic segmentation is broadly applied to several real-time applications including autonomous driving, medical imaging and robotics \cite{milioto2018real,paszke2016enet,salehi2018real,su2018real}.
Accurate CNN-based semantic segmentation requires larger neural networks with deeper and wider layers. These larger networks are therefore not suitable for edge computing devices as they are cumbersome and require substantial resources.

Down-sampling operations, such as pooling and convolutions with stride greater than one, can help decrease the latency of deeper neural networks, however they result in decreased pixel-level accuracy due to the lower resolutions at deeper levels.
Many recent approaches employ either encoder-decoder structure \cite{unet,badrinarayanan2017segnet,sun2018fishnet}, a two or multi-branch architecture \cite{poudel2019fastscnn,zhao2018icnet,yu2018bisenet} or dilated convolutions \cite{chen2014semantic,chen2017deeplab,chen2017rethinking,Zhao_2017} to recover spatial information.
While these real-time architectures perform appropriately on simple datasets, their performance is sub-optimal for complex datasets possessing more variability in terms of classes, sizes, and shapes.
Thus, there is a significant interest in designing CNN architectures that can perform well on complex datasets and, at the same time, are mobile enough to be of practical use in real-time applications.

In this paper, we propose a real-time general purpose semantic segmentation network, RGPNet, that performs well on complex scenarios. RGPNet is based on an asymmetric encoder-decoder structure with a new module called {\it adaptor} in the middle. The adaptor utilizes features at different abstraction levels from both the encoder and decoder to improve the feature refinement at a given level allowing the network to preserve deeper level features with higher spatial resolution. Furthermore, the adaptor enables a better gradient flow from deeper layers to shallower layers by adding short paths for the back-propagation. Since training an average deep learning model has a considerable carbon footprint \cite{strubell2019energy}, we reduce the training time by $60\%$ with negligible effect on performance by applying progressive resizing for training.

Our main contributions are as follows:
\begin{itemize}
    \item We propose RGPNet as a general real-time semantic segmentation architecture that obtains deep features with high resolution resulting in improved accuracy and lower latency in a single branch network. It performs competitively in complex environments.
    \item We introduce an adaptor module to capture multiple levels of abstraction to help in boundary refinement of segments. The adaptor also aids in gradient flow by adding short paths. 
    \item Towards green AI, we adopt progressive resizing technique during the training which leads to $60\%$ reduction in training time and the environmental impact. We combat aliasing effect in label map on lower resolutions by employing a modified label relaxation
    \item We report results on different datasets evaluated on single scale images. RGPNet achieves $80.9\%$, $69.2\%$, and $50.2\%$ mIoU with Resnet-101 backbone and $74.1\%$, $66.9\%$, and $41.7\%$ mIoU with Resnet-18 backbone on Cityscapes, CamVid and Mapillary, respectively. 
    \item For a $1024\times 2048$ resolution image, RGPNet(Resnet101) obtains 10.9 FPS in Pytorch and 15.5 FPS in TensorRT on NVIDIA RTX2080Ti GPU on the Cityscapes dataset whereas RGPNet(Resnet18) obtains 37.8 FPS and 47.2 FPS respectively under the same setting.
\end{itemize}

\section{Related Work}
\begin{figure*}[tb]
  \centering
  \includegraphics[width=0.9\textwidth, trim={0.5cm 4.5cm 0cm 5cm},clip]{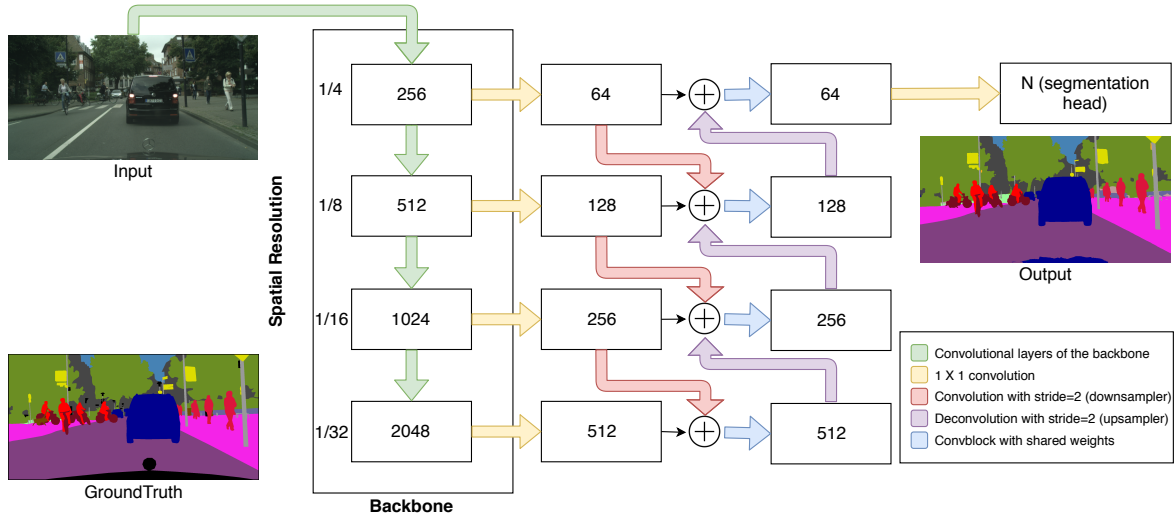}
  \caption{Network schematic diagram of the proposed architecture, RGPNet. Rectangular boxes depict tensor at a given level with number of channels mentioned as their labels. Color coded arrows represent the convolution operations indicated by the legend.}
  \label{fig:arch}
\end{figure*}

Semantic segmentation lies at the core of computer vision. With the advent of deep learning, Long \etal \cite{long2015fully} proposed the seminal fully convolutional network (FCN) with an end-to-end learning approach. However, FCN suffers from the loss of spatial details as it only utilizes high-level features from the last convolutional layer. Here, we summarize four widely used approaches which have been put forward that increase the feature resolution:

1) Context-based models:
To capture the contextual information at multiple scales, DeepLabV2 \cite{chen2014semantic} and DeeplabV3 \cite{chen2017deeplab} exploit multiple parallel atrous convolutions with different dilation rates, while PSPNet \cite{Zhao_2017} performs multi-scale spatial pooling operations. Although these methods encode rich contextual information, they can not capture boundary details effectively due to strided convolution or pooling operations \cite{deeplabv3plus2018}.

2) Encoder-decoder structure:
Several studies entail encode-decoder structure \cite{unet,badrinarayanan2017segnet,Pohlen_2017,zhuang2018shelfnet,li2018learning,ding2018context,fu2019stacked}. Encoder extracts global contextual information and decoder recovers the spatial information. Deeplabv3+ \cite{deeplabv3plus2018} utilizes an encoder to extracts rich contextual information in conjunction with a decoder to retrieve the missing object boundary details. However, implementation of dilated convolution at higher dilation rates is computationally intensive making them unsuitable for real-time applications.

3) Attention-based models:
Attention mechanisms, which help networks to focus on relevant information and ignore the irrelevant information, have been widely used in different tasks, and gained popularity to boost the performance of semantic segmentation. Wang \etal \cite{wang2018non} formalized self-attention by calculating the correlation matrix between each spatial point in the feature maps in video sequences.
To capture contextual information, DaNet \cite{fu2019dual} and OCNet \cite{yuan2018ocnet} apply a self-attention mechanism. DaNet has dual attention modules on position and channels to integrate local features with their respective global dependencies. OCNet, on the other hand, employs the self-attention mechanism to learn the object context map recording the similarities between all the pixels and the associated pixel. PSANet \cite{zhao2018psanet} learns to aggregate contextual information for each individual position via a predicted attention map. Attention based models, however, generally require expensive computation.

4) Multi-Branch models:
Another approach to preserve the spatial information is to employ two- or multi-branch approach. The deeper branches extract the contextual information by enlarging receptive fields and shallower branches retain the spatial details.
The parallel structure of these networks make them suitable for run time efficient implementations \cite{yu2018bisenet,zhao2018icnet,poudel2019_fastscnn, orsic2019defense_swiftnet}. However, they are mostly applicable to the relatively simpler datasets with fewer number of classes. On the other end, HRNet \cite{Sun19hrnet} proposed a model with fully connected links between output maps of different resolutions. This allows the network to generalize better due to multiple paths, acting as ensembles. However, without reduction of spatial dimensions of features, the computational overhead is very high and makes the model no longer feasible for real-time usage.

Building on these observations, we propose a real-time general purpose semantic segmentation architecture that obtains deep features with high resolution resulting in improved accuracy and lower latency in a single branch encoder-decoder network.

\section{Proposed Approach}
\subsection{Structure of RGPNet}
RGPNet's design is based on a light-weight asymmetric encoder-decoder structure for fast and efficient inference. It comprises of three components: an encoder which extracts high-level semantic features, a light asymmetric decoder, and an adaptor which links different stages of encoder and decoder.
The encoder decreases the resolution and increases the number of feature maps in the deeper layers, thus it extracts more abstract features in deeper layers with enlarged receptive fields. The decoder reconstructs the lost spatial information. The adaptor amalgamates the information from both encoder and decoder allowing the network to preserve and refine the information between multiple levels.

RGPNet architecture is depicted in Figure \ref{fig:arch}. In a given row of the diagram, all the tensors have the same spatial resolution with the number of channels mentioned in the scheme. Four level outputs from the encoder are extracted at different spatial resolutions $1/4$, $1/8$, $1/16$ and $1/32$ with 256, 512, 1024 and 2048 channels, respectively. The number of channels are reduced by a factor of four using $1\times1$ convolutions followed by batch norm and ReLU activation function at each level. These outputs are then passed through a decoder structure with adaptor in the middle. Finally, segmentation output is extracted from the largest resolution via $1\times1$ convolution to match the number of channels to segmentation categories.

\begin{figure}
  \includegraphics[width=0.43\textwidth]{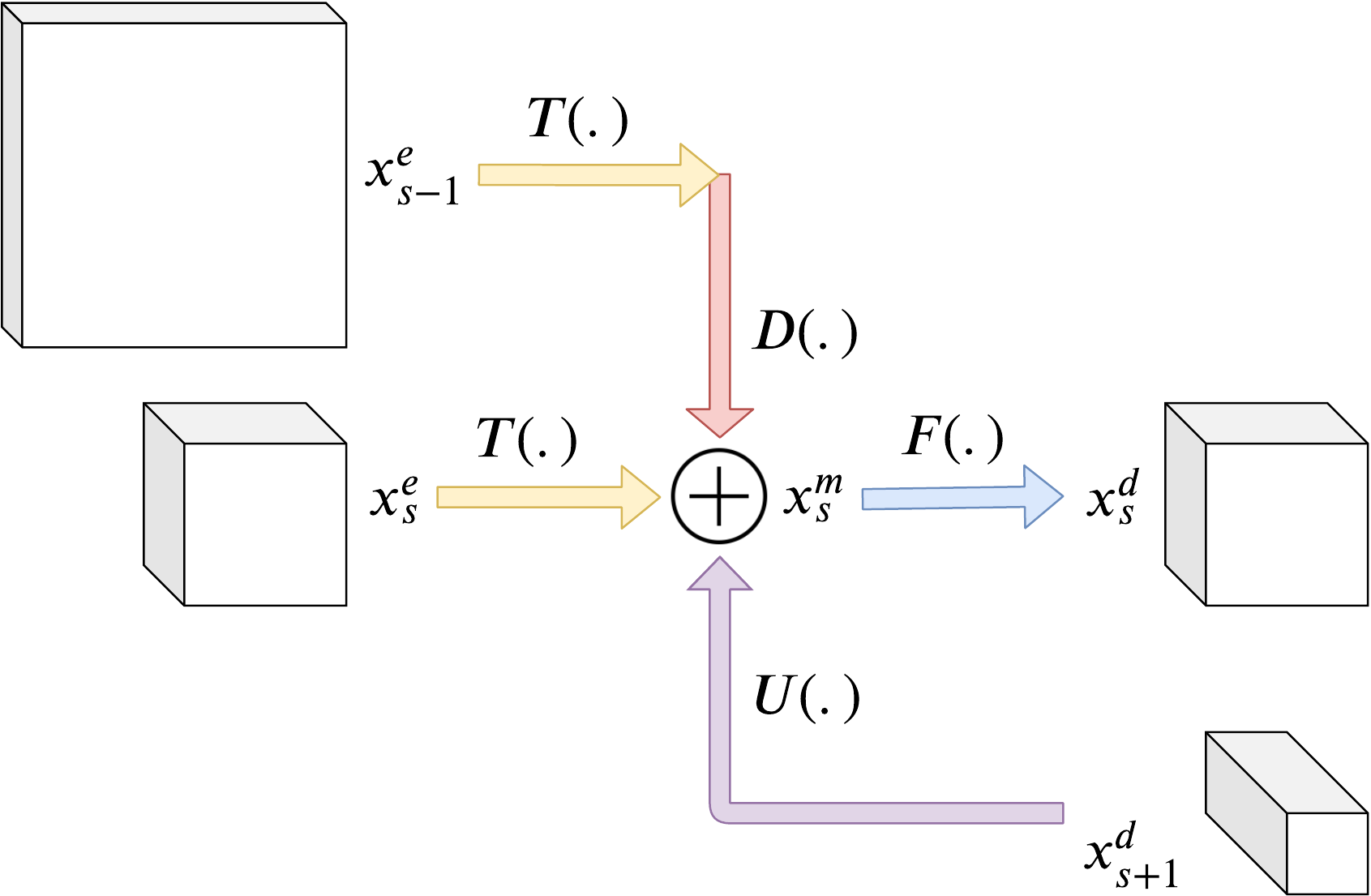}
  \caption{Adaptor module: the adaptor fuses information from multiple abstraction levels; $T(.)$, $D(.)$, and $U(.)$ denote the transfer, downsampling and upsampling functions, respectively. $F(.)$ is the decoder block with shared weights between layers.}
  \label{fig:functional_adaptor}
\end{figure}


{\bf Adaptor:}
Adaptor acts as a feature refinement module. The presence of an adaptor precludes the need of a symmetrical encoder-decoder structure. It aggregates the features from three different levels, and intermediates between encoder and decoder (Figure \ref{fig:functional_adaptor}). The adaptor function is as below:
\begin{equation}
    x_s^a=D(T(x_{s-1}^e))+T(x_s^e)+U(x_{s+1}^d)
\end{equation}
where superscripts $a$, $e$, and $d$ denote {\it adaptor}, {\it encoder}, and {\it decoder} respectively, $s$ represents the spatial level in the network. $D(.)$ and $U(.)$ are downsampling and upsampling functions. Downsampling is carried out by convolution with stride 2 and upsampling is carried out by deconvolution with stride 2 matching spatial resolution as well as the number of channels in the current level. $T(.)$ is a transfer function that reduces the number of output channels from an encoder block and transfers them to the adaptor:
\begin{equation}
    T(x_s^e)=\sigma(\omega_s^a\otimes x_s^e+b_s^a)
\end{equation}
where $\omega$ and $b$ are the weight matrix and bias vector, $\otimes$ denotes the convolution operation, and $\sigma$ denotes the activation function.
The decoder contains a modified basic residual block, $F$, where we use shared weights within the block. The decoder function is as follows:
\begin{equation}
 x_s^d = F(x_s^m;\omega_s^d)
\end{equation}

The adaptor has a number of advantages. First, the adaptor aggregates features from different contextual and spatial levels. Second, it facilitates the flow of gradients from deeper layers to shallower layers by introducing a shorter path. Third, the adaptor allows for utilizing asymmetric design with light-weight decoder. This results in fewer convolution layers, further boosting the flow of gradients. The adaptor, therefore, makes the network suitable for real-time applications as it provides rich semantic information while preserving the spatial information.

\subsection{Progressive Resizing with Label Relaxations}
Progressive resizing is a technique commonly used in classification to reduce the training time. The training starts with smaller image sizes followed by a progressive increase of size until the final stage of the training is conducted using the original image size. For instance, this technique can theoretically speed up the training time by $16$ times per epoch if the image dimensions are decreased by $1/4$, and correspondingly the batch size is increased by a factor of $16$ in a single iteration. 

However, applying progressive resizing for semantic segmentation is more challenging as the resizing method should be applied to images and their corresponding label maps. Bi-linear or bi-cubic interpolation cannot be applied to label maps as they exist in integer space and these methods will result in float values for labels. On the other hand, nearest-neighbor interpolation for resizing introduces noise in the label maps around the borders of the objects due to aliasing. To reduce the effects of boundary artifacts in progressive resizing for label maps, inspired by Zhu \etal \cite{zhu2018improving}, we propose an optimized variant of the label relaxation method.

In cross-entropy loss function, the negative log-likelihood of softmax probability for a given label is maximized. In contrast, label-relaxation is a custom loss function where the negative log-likelihood of softmax probabilities for a given label as well as bordering pixel labels is maximized.
This is established by taking the sum of softmax probabilities mentioned, before applying the negative log-likelihood. We identify the border pixels as those which have more than one unique label in the window with kernel size $k$ centered on it. The loss at a given border pixel can be calculated as follows where N is a set of border labels:

\begin{equation}
    L_{boundary} = -\log \sum_{C \in N} P(C)
\end{equation}

To apply label relaxation efficiently, first, one-hot labels are created from the label map followed by max-pool operation with stride $1$. This effectively dilates each one-hot label channel transforming it into multi-hot labels along the borders which enables optimized selection of border pixels along with their corresponding labels. Border pixels are usually in minority; for instance, in the Cityscapes dataset, on average, only $2.4\%$ of total pixels are border pixels. This loss function is only applied to border pixels whereas normal cross-entropy loss is applied to the rest of the pixels.

\begin{figure*}[tb]
 \centering
 \includegraphics[width=0.98\textwidth]{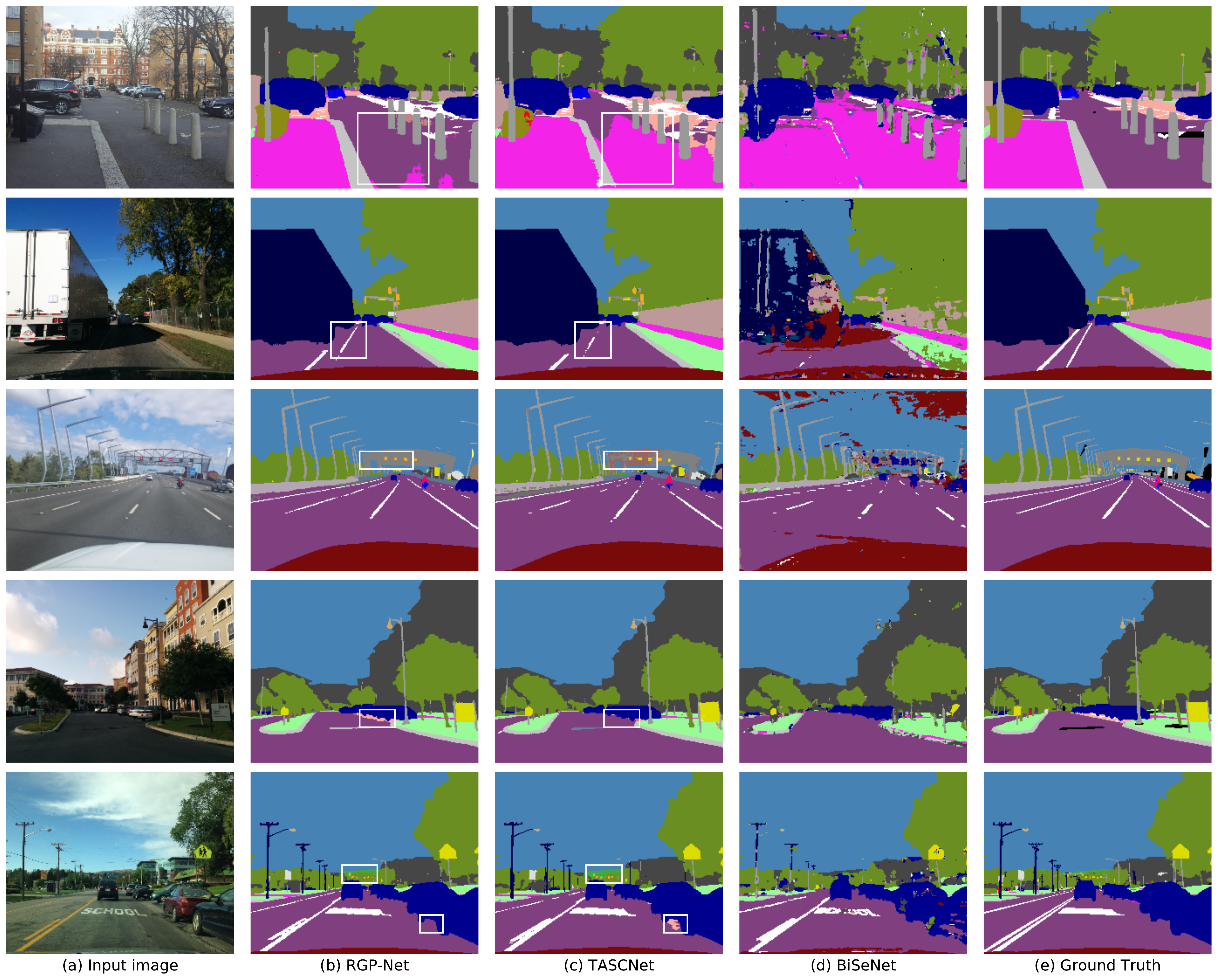}
 \caption{Semantic segmentation results on Mapillary Vistas validation set. The columns correspond to input image, the output of RGPNet, the output of TASCNet, the output of BiSeNet, and the ground-truth annotation. For all methods R101 is used as the backbone. RGPNet mainly improves the results on road and road-related objects' pixels. Best viewed in color and with digital zoom.}
 \label{fig:vis_mapillary}
\end{figure*}

\section{Experimental Results}
We conduct experiments on Mapillary \cite{neuhold2017mapillary} as a highly complex dataset, CamVid \cite{brostow2009semantic} and Cityscapes \cite{Cityscapes_Cordts_2016} as moderately complex datasets.

{\bf Mapillary} consists of $20,000$ high-resolution street-level images taken from many different locations around the globe and under varying conditions annotated for $65$ categories. The dataset is split up in a training set of $18,000$ images and a validation set of $2,000$ images. 

{\bf CamVid}
 consists of 701 low-resolution images in 11 classes which are divided into 376/101/233 image sets for training, validation and testing, respectively. Here, we use the same experimental setup as SegNet \cite{badrinarayanan2017segnet}: $352 \times 480$ image resolution for training and inference, 477 images for training and validation, and 233 image as test set.

{\bf Cityscapes} contains diverse street level images from $50$ different cities. It contains $30$ classes and only $19$ classes of them are used for semantic segmentation evaluation. The dataset contains $5000$ high quality pixel-level finely annotated images and $20000$ coarsely annotated images. The finely annotated $5000$ images are divided into $2975/500/1525$ image sets for training, validation and testing. We do not use coarsely annotated data in our experiments.

\begin{table*}[htb]
\caption{CamVid test set results calculated on $352 \times 480$ image resolution. The inference times are calculated on a single NVIDIA TitanV GPU with a single-image batch size. \label{table_camvid}}
\centering
\resizebox{\textwidth}{!}{%
\begin{tabular}{|l|cc|ccccccccccc|cc|}
\hline
Model(backbone) & \rotatebox{90}{Params(M)} & \rotatebox{90}{FPS} & \rotatebox{90}{Building} & \rotatebox{90}{Tree} & \rotatebox{90}{Sky} & \rotatebox{90}{Car} & \rotatebox{90}{Sign} & \rotatebox{90}{Road} & \rotatebox{90}{Pedestrian} & \rotatebox{90}{Fence} & \rotatebox{90}{Pole} & \rotatebox{90}{Sidewalk} & \rotatebox{90}{Cyclist} & \rotatebox{90}{mIoU(\%)} & \rotatebox{90}{Pixel Acc.} \\ \hline
SegNet & 29.5 & 63.0 & 68.7 & 52.0 & 87.0 & 58.5 & 13.4 & 86.2 & 25.3 & 17.9 & 16.0 & 60.5 & 24.8 & 46.4 & 62.5 \\
FCN8 & 135 & 47.6 & 77.8 & 71.0 & 88.7 & 76.1 & 32.7 & 91.2 & 41.7 & 24.4 & 19.9 & 72.7 & 31.0 & 57.0 & 88.0 \\
FC-DenseNet56 & \bf 1.4 & 38.2 & 77.6 & 72.0 & 92.4 & 73.2 & 31.8 & 92.8 & 37.9 & 26.2 & 32.6 & 79.9 & 31.1 & 58.9 & 88.9 \\
FC-DenseNet103 & 9.4 & 20.4 & 83.0 & \bf 77.3 & \bf 93.0 & 77.3 & 43.9 & 94.5 & 59.6 & 37.1 & 37.8 & \bf 82.2 & 50.5 & 66.9 & \bf 91.5 \\
FC-HarDNet68 & \bf 1.4 & 75.2 & 80.8 & 74.4 & 92.7 & 76.1 & 40.6 & 93.3 & 47.9 & 29.3 & 33.3 & 78.3 & 45.7 & 62.9 & 90.2 \\
FC-HarDNet84 & 8.4 & 34.8 & 81.4 & 76.2 & 92.9 & 78.3 & 48.9 & \bf 94.6 & 61.9 & 37.9 & 38.2 & 80.5 & 54.0 & 67.7 & 91.1 \\ \hline
\bf RGPNet(R18) & 17.7 & \bf 190 & 82.6 & 75.5 & 91.2 & 85.1 & 54.3 & 94.1 & 61.5 & 50.4 & 36.8 & \bf 82.2 & 59.8 & 66.9 & 90.2 \\
\bf RGPNet(R101) & 50.1 &  68.2 & \bf 85.8 & \bf 77.3 & 91.2 & \bf 87.0 & \bf 62.5 & 90.6 & \bf 67.6 & \bf 51.4 & \bf 46.8 & 70.7 & \bf 67.2 & \bf 69.2 & 89.9 \\ \hline
\end{tabular}%
}
\end{table*}

We implement the RGPNet based on PyTorch framework \cite{paszke2017automatic}. For training on both datasets, we employ a polynomial learning rate policy where the initial learning rate is multiplied by $(1 - iter/total\_iter )^{0.9}$ after each iteration. The base learning rate is set to $1 \times 10^{-3}$. Momentum and weight decay coefficients are set to $0.9$ and $1 \times 10^{-4}$, respectively. We train our model with synchronized batch-norm implementation provided by Zhang \etal \cite{Zhang_2018_CVPR}. Batch size is kept at $12$ and trained on two Tesla V100 GPUs. For data augmentation, we apply random cropping and re-scaling with $1024$ as crop-size. Image base size is $1536$ for Mapillary and $2048$ for Cityscapes. Re-scaling is done from range of $0.5$ to $2.0$ respectively followed by random left-right flipping during training.

As a loss function, we use cross entropy with online hard example mining (OHEM) \cite{wu2016high,yuan2018ocnet}. OHEM only keeps the sample pixels which are hard for the model to predict in a given iteration. The hard sample pixels are determined by probability threshold $\theta$ for the corresponding target class, thus the pixels below the threshold are preserved in the training. To have enough representative of each class in the mini batch, the minimal pixel ratio $M$ is applied. In our experiments, we set $\theta=0.6$ and $M=5 \times 10^3$.

\subsection{Results on Mapillary}
In this section, we evaluate and compare overall performance of RGPNet with other real-time semantic segmentation methods (BiSeNet \cite{yu2018bisenet}, TASCNet \cite{li2018learning}, and ShelfNet \cite{zhuang2018shelfnet}) on Mapillary validation set. we use different feature extractor backbones ResNet \cite{He_2016_Resnet} (R101, R50 and R18), Wide-Resnet \cite{Wider_or_Deeper_Wu_2019} (WRN38), and HarDNet \cite{chao2019hardnet} (HarDNet39D). 

Table \ref{table_mapillary} compares speed (FPS), mIoU and number of parameters on these methods on 32-bit precision computation. RGPNet(R101) achieves $50.2\%$ mIoU which outperforms TASCNet and ShelfNet with a significant margin and lower latency. Although RGPNet(R101) has more parameters than the TASCNet(R101), both speed and mIoU are considerably higher. However, BiSeNet demonstrates poor performance on Mapillary resulting in the lowest mIoU. Using TensorRT, RGPNet (R101 as the encoder) speeds up to $61.9$ FPS on full image resolution (Table \ref{table_tensorrt}). Our method also achieves impressive results with a lighter encoder (R18 or HarDNet39D) surpassing BiSeNet with a heavy backbone (R101) significantly, $41.7\%$ vs $20.4\%$ mIoU and 54.4 vs 15.5 FPS. Finally, Figure \ref{fig:vis_mapillary} shows some qualitative results obtained by our model compared to TASCNet and BiSeNet.

\begin{table}[htb]
\centering
\caption{Mapillary Vistas validation set. Inference speed is calculated on $1024 \times 2048$ image resolution.}
\label{table_mapillary}
\begin{tabular}{|l|ccc|}
\hline
Model(backbone) & FPS & mIoU(\%) & Params(M) \\ \hline
BiSeNet(R101)       & 9.27          & 20.4          & 50.1 \\
TASCNet(R50)        & 11.9          & 46.4          & 32.8 \\
TASCNet(R101)       & 8.84          & 48.8          & 51.8 \\
ShelfNet(R101)      & 9.11          & 49.2          & 57.7 \\
\bf RGPNet(R101) & \bf 10.8    & \bf 50.2    & \bf 52.2 \\
RGPNetB(WRN38) & 3.37  &  53.1    & 215 \\\hline
RGPNet(HarDNet39D) & 34.7 & 42.5 & 9.4 \\
\bfseries RGPNet(R18) & \bfseries 35.7 & \bfseries 41.7 & \bfseries 17.8 \\\hline
\end{tabular}
\end{table}

\subsection{Results on Camvid}
In Table \ref{table_camvid}, we compare overall performance of RGPNet with other real-time semantic segmentation methods (SegNet, FCN \cite{long2015fully}, FC-DenseNet \cite{jegou2017one}, and FC-HarDNet \cite{chao2019hardnet}) on CamVid test set. We find that RGPNet with R18 and R101 backbones obtain $66.9\%$ and $69.2\%$ mIoU with $190$ and $68.2$ FPS. RGPNet achieves significant increase in mIoU for Car, Traffic Sign, Pole, and Cyclist categories. Overall we observe that our model outperforms the state-of-the-art real-time segmentation models.

\subsection{Results on Cityscapes}
Table \ref{table_cityscape_higher_Accuracy} shows the comparison between our RGPNet and state-of-the-art real-time (BiSeNet, ICNet \cite{zhao2018icnet}, FastSCNN \cite{poudel2019fastscnn}, and ContextNet \cite{poudel2018contextnet}) and offline (HRNet \cite{Sun19hrnet} and Deeplabv3 \cite{deeplabv3plus2018}) semantic segmentation methods on Cityscapes validation dataset. RGPNet achieves $74.1\%$ mIoU which is slightly lower than BiSeNet $74.8\%$ mIoU. ICNet, ContextNet and FastSCNN achieve lower mIoU. Compared to the heavy offline segmentation methods, RGPNet(R101) not only is the fastest, but also outperforms Deeplabv3, BiSeNet (R101) and is comparable to HRNet.

We, therefore, conclude that RGPNet is a real-time general purpose semantic segmentation model that performs competitively in a wide spectrum of datasets compared to the state-of-the-art semantic segmentation networks designed for specific datasets.

\begin{table}[tb]
\caption{Cityscapes validation set result on $1024 \times 2048$ image. Numbers with * are taken from respective paper. SS and MS denote single-scale and multi-scale. OOM stands for out-of-memory error. Numbers with $\dagger$ are computed in TensorFlow framework with our in-house implementations which are better than originally reported in respective paper.}
\label{table_cityscape_higher_Accuracy}
\centering
\begin{tabular}{|ll|ccc|}
\hline
\multicolumn{2}{|c|}{Model} & \multicolumn{2}{c}{mIoU (\%)} & \multirow{2}{*}{FPS} \\\cline{1-4}
Backbone & Head & SS & MS & \\ \hline
R18 & BiSeNet & 74.8* & 78.6* & 40.4 \\ 
PSPNet50 & ICNet & 67.7* & - & 40.6$\dagger$ \\ 
N/A & FastSCNN & 68.1 & - & 43.5$\dagger$ \\ 
N/A & ContextNet & 60.6 & - & 37.9$\dagger$ \\ 
R18 & SwiftNet &  75.4* & - & 56.3 \\ 

\bf R18 & \bf RGPNet & \bf 74.1 & \bf 76.4 & \bf 37.8 \\ \hline 
W48 & HRNet & 81.1* & - & OOM \\
R101(OS-8) & Deeplabv3 & 77.82* & 79.30* & 2.48 \\ 
R101 & BiSeNet & - & 80.3* & 10.4 \\
\bf R101 & \bf RGPNet & \bf 80.9 & \bf 81.9 & \bf 10.9 \\ \hline 
\end{tabular} %
\end{table}

\subsection{Progressive resizing with label relaxation}
\begin{figure}[tb]
  \centering
  \includegraphics[width=1\columnwidth, trim={0cm 0cm 0cm .7cm},clip]{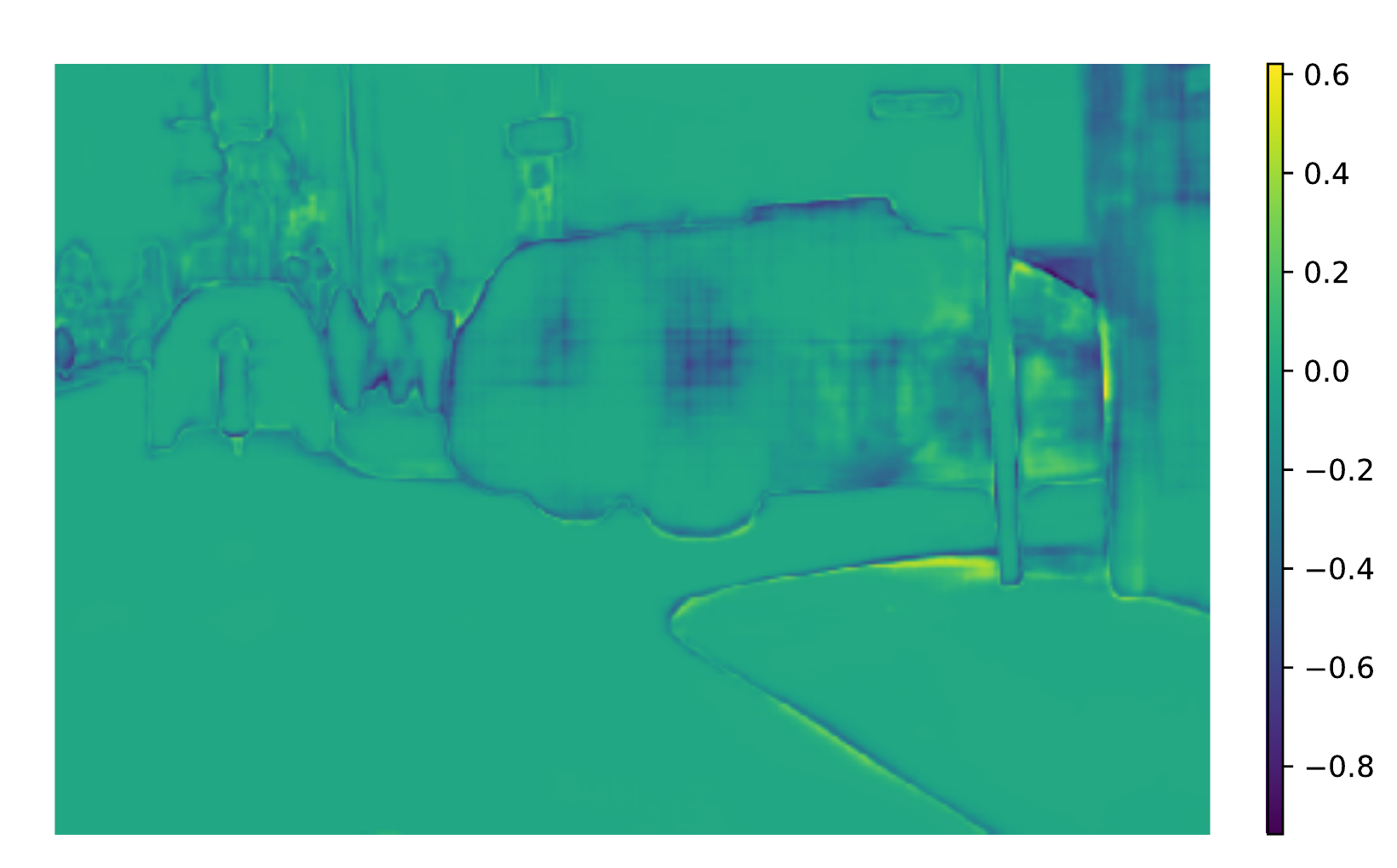}
  \caption{Heatmap of difference in entropy between label relaxation and without label relaxation based trained model evaluated on a sample image from validation set. On boundaries of objects, models trained with label relaxation are more confident about the label and hence have lower entropy (blue shades).}
  \label{fig:vis_entropy}
\end{figure}

Here, we compare the result of progressive resizing training with and without label relaxation. In these experiments for the first 100 epochs, the input images are resized by a factor of $1/4$ both in width and height. At the $100^{th}$ epoch, the image resize factor is set to $1/2$ and, at the $130^{th}$ epoch, full-sized images are used. To analyze the effect of label relaxation in progressive resizing technique, we illustrate the difference in entropy between two setups (progressive resizing with and without label relaxation). Figure \ref{fig:vis_entropy} shows that the model trained with label relaxation is more confident in the predictions around the object boundaries. 



\paragraph{Green AI}
To examine the energy efficiency, we run the experiments with and without progressive resizing training technique and label relaxation on a single GPU for 15 epochs. In the standard training experiment, we use a full size Cityscapes image $1024 \times 2048$. In the progressive resizing training experiment, we start with $1/4$ of image size and then scale up by a factor of 2 at the $10^{th}$ and the $13^{th}$ epochs. The speedup factor can theoretically be calculated as $1/16 \times 9/15 + 1/4 \times 3/15 + 3/15 = 0.2875$. Table \ref{table_progressive} shows that the training time reduced from 109 minutes to 32 minutes, close to the speedup expected from theoretical calculation. Note that, inclusion of our optimized label relaxation causes a minute increase in the energy consumption (less than 10 KJ) and time (4 minutes more). The energy consumed by GPU decreases by an approximate factor of 4 when compared to full scale experiment with slight drop in the performance. Towards green AI, as a result of remarkable gain in energy efficiency, we therefore suggest adopting progressive resizing technique with label relaxation for training a semantic segmentation network.

\begin{table}[htb]
\caption{Progressive resizing result on energy efficiency. PR and LR stand for progressive resizing and label relaxation, respectively. mIoU reported here are from the complete experiment.}
\label{table_progressive}
\centering
\resizebox{\columnwidth}{!}{%
\begin{tabular}{|l|ccc|}
\hline
Training Scheme & Energy(KJ) & Time & mIoU(\%)\\ \hline
PR w/o LR & 203 & 27m 37s &  78.3  \\  
PR with LR & 212 & 31m 43s & 78.8 \\  
Full scale & 873 & 108m 44s & 80.9 \\ 
\hline
\end{tabular}}
\end{table}

\begin{figure*}[htb]
 \centering
 \includegraphics[width=.9\textwidth, trim={8cm 3cm 6cm 3cm},clip]{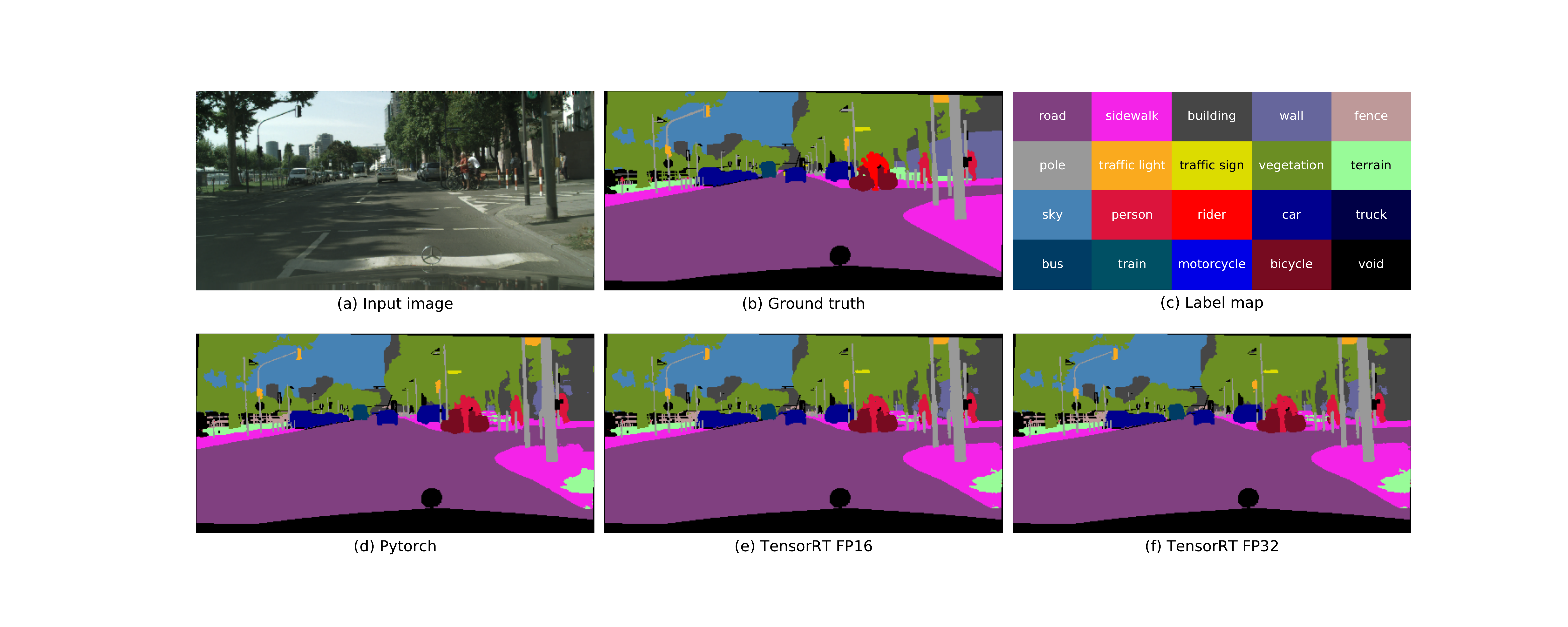}
 \caption{Results obtained by RGPNet on Cityscapes validation set on $1024 \times 2048$ image resolution. Top row: input image, ground-truth annotation, and label maps. Bottom row: the output of PyTorch model, TensorRT FP16 model, and TensorRT FP32 model. The results show that optimization on TensorRT on half and full precision floating point format does not affect the qualitative outputs.}
 \label{fig:vis_seg}
\end{figure*}

\begin{table*}[htb]
\centering
\caption{RGPNet inference speed (FPS) using TensorRT on Nvidia RTX2080Ti and Xavier evaluated on half and full resolution images from Cityscapes dataset. Comparison between floating point 16 bit (FP16) and 32 bit (FP32) computation is also shown.\label{table_tensorrt}}
\begin{tabular}{|l|rrrr|rrrr|}
\hline
 \multirow{3}{*}{Backbone} & \multicolumn{4}{c|}{Nvidia RTX2080Ti} & \multicolumn{4}{c|}{Xavier} \\ \cline{2-9} 
 & \multicolumn{2}{c|}{$512 \times 1024$} & \multicolumn{2}{c|}{$1024 \times 2048$} &
 \multicolumn{2}{c|}{$512 \times 1024$} & \multicolumn{2}{c|}{$1024 \times 2048$} \\ \cline{2-9} 
 & \multicolumn{1}{l|}{FP16} & \multicolumn{1}{l|}{FP32} & \multicolumn{1}{l|}{FP16} &
 \multicolumn{1}{l|}{FP32} & \multicolumn{1}{l|}{FP16} & \multicolumn{1}{l|}{FP32} &
 \multicolumn{1}{l|}{FP16} & \multicolumn{1}{l|}{FP32} \\ \hline
 R18 & 430.2 & 180.9 & 153.4 & 47.2 & 78.45 & 24.6 & 20.8 & 6.17 \\
 R50 & 265.7 & 88.8 & 87.2 & 24.3 & 44.6 & 12.6 & 11.7 & 3.17 \\
 R101 & 176.9 & 58.5 & 61.9 & 15.5 & 30.3 & 8.14 & 7.89 & 2.05 \\
\hline
\end{tabular}
\end{table*}

\subsection{Ablation study}
In this section, we perform an empirical evaluation on the structure of the adaptor module in our design.
We show the significance of the downsampling layers which provides information from a higher resolution of encoder to adaptor. Table \ref{table_ablation_mapillary} shows that the performance of our model significantly drops from $50.2\%$ to $46.8\%$ on Mapillary validation set when the downsampling layers are removed. This indicates that the specific design of adaptor has an important role in feature preserving and refinement in our model. We show similar effect in in Table \ref{table_ablation} which showcases results on Cityscapes dataset where adding downsampling layers result in boost in mIoU with Resnet101 backbone. 

\begin{table}[htb]
\caption{Ablation study on Mapillary validation set: it highlights the effect of downsampling layers which are shown in red in Figure \ref{fig:arch} from adaptor. MS+F stands for multi-scale evaluation with left/right image flip.}
\label{table_ablation_mapillary}
\centering
\begin{tabular}{|l|c|c|}
\hline
Method & DS & MS+F mIoU(\%) \\ \hline
RGPNet(R101) & & 46.8 \\ 
RGPNet(R101) & \checkmark & 50.2 \\ \hline
\end{tabular}
\end{table}

We perform an ablation study on components of our training framework on Cityscapes dataset. These techniques are namely; cross entropy (CE), addition of downsampling layers (DS), cross entropy with online hard example mining (OHEM) and using pretrain weights from Mapillary (PM). 
For the last technique mentioned, we adopt a pretrained model on Mapillary dataset by sorting the last layer weights according to mapping between Maplillary and Cityscapes categories. This results in more than $4.2\%$ and $2.4\%$ boost in mIoU with Resnet101 and Resnet18 backbones respectively while evaluating on multiple scales.
As illustrated in Table~\ref{table_ablation}, all these components contribute significantly in improving the final performance.

\begin{table}[htb]
\caption{Ablation study on Cityscapes validation set with RGPNet on $1024 \times 2048$ image resolution. CE, DS, OHEM, PM denote cross-entropy loss, down-sampling connections, online hard example mining loss, and pretrained model on Mapillary, respectively. MS+F stands for multi-scale evaluation with left/right image flip.}
\label{table_ablation}
\centering
\resizebox{\columnwidth}{!}{%
\begin{tabular}{|cccc|cc|cc|}
\hline
\multirow{2}{*}{CE} & \multirow{2}{*}{DS} &  \multirow{2}{*}{OHEM} &  \multirow{2}{*}{PM} & \multicolumn{2}{|c|}{R101} & \multicolumn{2}{|c|}{R18} \\ \cline{5-8}
 &  &  &  & SS & MS+F & SS & MS+F \\ \hline
 \checkmark & & & & 73.0 & 74.6 & 69.1 & 71.2 \\
 \checkmark & \checkmark & & & 73.1 & 75.5 & 69.0 & 71.3\\
 \checkmark & \checkmark & \checkmark & & 76.5 &  77.7 & 71.9 & 74.0 \\
 \checkmark & \checkmark & \checkmark & \checkmark & 80.9 &  81.9 & 74.1 &  76.4\\ \hline
\end{tabular}}
\end{table}

\subsection{TensorRT}
We use TensorRT for RGPNet and evaluate on Nvidia RTX2080Ti and Xavier. RGPNet obtains $79.26\%$ and $79.25\%$ mIoU on Cityscpaes validation with half and full precision floating point format, respectively. The inference speed results for different backbones, two input resolutions using 16-bit and 32-bit floating point numbers are reported in Table \ref{table_tensorrt}. RGPNet(R18) using TensorRT on full input resolution leads to a significant increase in speed from 37.8 FPS to 153.4 FPS with 16-bit floating point operations. The speed up with FP16 compared to FP32 is noticeable for all backbones, and two different input resolutions. The results suggest that RGPNet can run high speed on edge computing devices with little or negligible drop in accuracy. A real-world example is provided in Figure \ref{fig:vis_seg}.

\section{Conclusion}
In this paper, we proposed a real-time general purpose semantic segmentation network, RGPNet. It incorporates an adaptor module that aggregates features from different abstraction levels and coordinates between encoder and decoder resulting in a better gradient flow. Our conceptually simple yet effective model achieves efficient inference speed and accuracy on resource constrained devices in a wide spectrum of complex domains.
By employing an optimized progressive resizing training scheme, we reduced training time by more than half with a small drop in performance, thereby substantially decreasing the carbon footprint.
Furthermore, our experiments demonstrate that RGPNet can generate segmentation results in real-time with comparable accuracy to the state-of-the-art non real-time models. 
This optimal balance of speed and accuracy makes our model suitable for real-time applications such as autonomous driving where the environment is highly dynamic due to the presence of high variability in real world scenarios.

{\small
\bibliographystyle{ieee_fullname}
\bibliography{egbib_imported}
}

\end{document}